# A Novel Attention-Based Network for Fast Salient Object Detection

Bin Zhang[1], Yang Wu[2], Xiaojing Zhang[3] and Ming Ma[*]

College of Computer Science and Engineering, Inner Mongolian University, China, Hohhot
csmaming@imu.edu.cn

## ABSTRACT

*In the current salient object detection network, the most popular method is using U-shape structure. However, the massive number of parameters leads to more consumption of computing and storage resources which are not feasible to deploy on the limited memory device. Some others shallow layer network will not maintain the same accuracy compared with U-shape structure and the deep network structure with more parameters will not converge to a global minimum loss with great speed. To overcome all of these disadvantages, we proposed a new deep convolution network architecture with three contributions: (1) using smaller convolution neural networks (CNNs) to compress the model in our improved salient object features compression and reinforcement extraction module (ISFCREM) to reduce parameters of the model. (2) introducing channel attention mechanism in ISFCREM to weigh different channels for improving the ability of feature representation. (3) applying a new optimizer to accumulate the long-term gradient information during training to adaptively tune the learning rate. The results demonstrate that the proposed method can compress the model to 1/3 of the original size nearly without losing the accuracy and converging faster and more smoothly on six widely used datasets of salient object detection compared with the others models. Our code is published in https://gitee.com/binzhangbinzhangbin/code-a-novel-attention-based-network-for-fast-salient-object-detection.git*

## KEYWORDS

*Salient Object Detection, Optimization Strategy, Deep Learning, Model Compression, Vision Attention*

## 1. INTRODUCTION

Salient object detection plays an important role in many computer vision tasks, such as computer vision tracking [1], content aware image processing [2], medical segmentation, robot vision navigation [3] and so on. Most of traditional detection methods [4,5,6,7,8,9,10,11] extract handle-crafted features to capture low-level visual features such as color, intensity and orientation. Traditional models are also dependent on the manually designed regional salient descriptors. However, low-level semantic information which means local contrast operator has a limited spatial neighborhood and large high-level feature information can be easily missed out. With the advent of convolution neural network, it can not only capture the low-level semantic information but also extract the high-level semantic information, which greatly improves the accuracy of significance by fusing the context details features of multi-scale space.

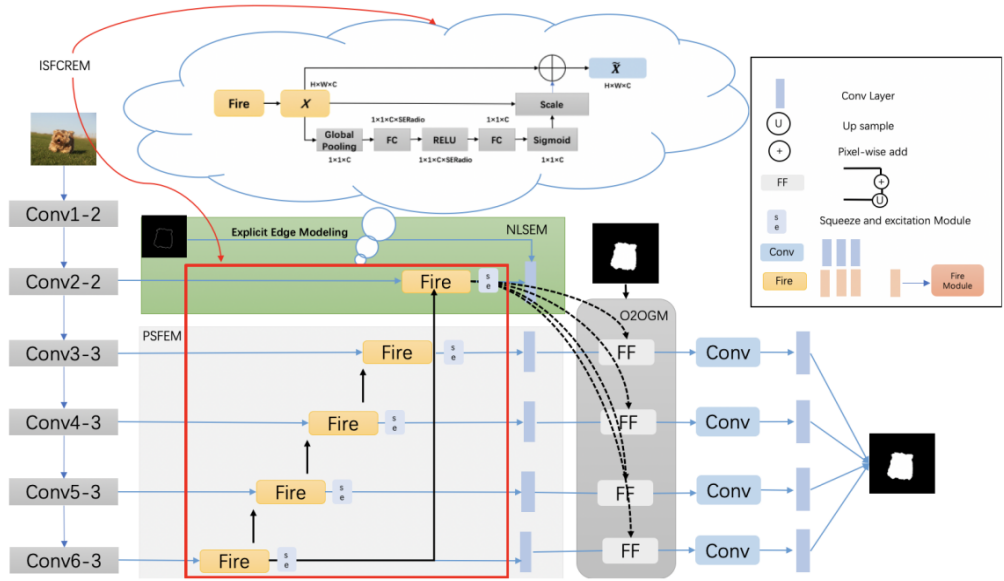

Figure 1. The pipeline of the proposed approach. Based on the Edge Guidance Network we replace the convolution layers of the decoder of U-shape structure with fire module. Follows we use squeeze-and-excitation module to focus more attention on different channel. PSFEM: progressive salient object features extraction module. NLSEM: non-local salient edge features extraction module. O2OGM: one-to-one guidance module. FF: feature fusion. ISFCREM: improved salient object features compression and reinforcement extraction module.

Methods [12,13,14] pointed out that the shape of CNN pyramid structure usually has a larger space and contains detailed low-order information in the shallow stage, while the deeper stage will contain more high-order information which can accurately highlight the location information of the positioning target. The U-shape structure [15,16] is used in the salient object detection framework [17,18,19] designed by fusing high-order and low-order information, which has caught global low-high level information by uniting different phase features. It constructs the feature fusion path from top to bottom in the classification network which means the low-level semantic information can better fuse the high-level semantic information so as to build a rich feature map.

Recently, based on the U-shape structure designed a new network that is EGNet [20] for salient objection detection which benefit from the rich edge information and location information in salient edge features. The EGNet can help locate salient objects more efficiently. Most of images utilize image blocks as input then using context information and end-to-end deep network structure to output salient map based on each pixel in the image region. After fusing edge information, the local edge information of the object is more prominent and it is used to refine the high-order semantic information. At the same time, the complete consistency between the edge information and the object information is highlighted. However, we found that the original convolution network can be compressed which received the influence of SqueezeNet [21] architecture when fusing the edge information and the salient map information of the context. We changed the convolution design on the premise of nearly keeping the receptive field unchanged and used the compressed convolution to reduce the parameters of the model. We also find that on the basis of the U-shape structure paying attention to the relationship between the channel can automatically learn the importance of different features according to different channel features so introducing channel attention mechanism is to make up for the loss of the details of the feature map caused by the compression of the model.

We focus the balanced compensation between the model compression of net and the precise of feature maps to save storage space on the premise of almost keeping the accuracy of the original model unchanged. Then we introduce the attention mechanism to focus on the weight attention of the different model channels in order to focus more on salient feature maps and make up for the loss caused by compression. In summary, this paper makes three major contributions:

- We design a new model structure with reduced model size which using smaller CNN architectures are more feasible to deploy on FPGAs and other hardware with limited memory and require less communication across servers during distributed training.

- We introduce the channel attention mechanism to increase the feature focus degree between different channels when the high-order semantic information and low-order semantic information are fused in the up sampling to improve the focus degree of salient map detection offsetting the loss caused by compression.

- We use a new optimization strategy for fast convergence instead of traditional optimization strategy which would possibly lead to local minimums only half of the training time compared with original model.

On the whole, we introduced the related works of the current development method in Sec 2. Next we proposed our network structure and concisely introduced the related modules respectively in Sec 3. Sec 4 is the analysis of results compared with others experiments. Finally we concluded our work in Sec 5.

## 2. RELATED WORKS

Recently, benefiting from the powerful feature extraction capability of CNNs, most of the traditional salient detection methods based on hand-crafted features [22] have been gradually surpassed.

In recent years, the commonly used salient object detection methods based on deep neural network are mostly optimized from the following three aspects: the method based on boundary enhancement, the method based on semantic enhancement, and the method based on boundary with semantic enhancement. Boundary based enhancement is to obtain more boundary information by enhancing the low-level features of depth features to better locate the significant target boundary. Typical algorithms include ELD [23] algorithm proposed by Tai et al., KSR [24] algorithm proposed by Wang et al., DCL [25] algorithm proposed by Li et al., and DSS algorithm proposed by Hou et al. Semantic enhancement is to obtain rich semantic information from high-level features to better locate the salient target and make the salient target more prominent. Typical algorithms include R-FCN [26] algorithm proposed by Dai et al., CPD [27] algorithm proposed by Wu et al., and PoolNet [28] algorithm proposed by Liu et al. These methods can accurately locate the position of salient object. However, the semantics contained in high-level features are enhanced but sometimes the boundary is blurred or multiple salient object is overlaid. It is easy to cause ambiguity of salient object while only semantic enhancement or boundary enhancement. In order to overcome the shortcomings of the two questions, some researches enhance the boundary and semantics at the same time which obtain good salient object information and contour information to improve the performance of salient object detection. Typical algorithms include Amulet algorithm proposed by Zhang et al. And BDMPM [29] algorithm proposed by Zhang et al.

Compared with the above network structure design, the popular edge information fusion strategy of EGNet makes up for the salient graph information. Through an end-to-end model design, we can better capture the salient graph information. We explore the model structure design and use the model compression strategy to reduce the parameters improving the model inference speed while maintaining the same sensing field using smaller CNNs. We introduce the channel attention mechanism to focus on the focusing degree between different map channels to enhance the correlation between channels to better make up for the loss caused by using model compression and use a new optimization strategy to converge the global loss minimum more quickly.

## 3. NETWORK ARCHITECTURE

The overall architecture is shown in Figure 1. In this section, we begin by describing the compression module in Sec. 3.1, then introduce the channel attention mechanism module in Sec. 3.2, introduce our improved salient object features compression and reinforcement extraction module (ISFCREM) in Sec. 3.3, and finally introduce the new AdaX optimizer in Sec. 3.4.

We also use the VGG [30] network as backbone, followed by the original EGNet we obtain six side features extracting Conv1-2, Conv2-2, Conv3-3, Conv4-3, Conv5-3, Conv6-3. Because the Conv1-2 is too close to the input and the receptive field is too small, we leverage the widely used architecture U-shape from Conv2-2 to Conv6-3 to generate more robust salient object features. But we use fire module instead of original up sampling layers to compress the model. Before adapt a single convolution layer to convert the feature maps to the single-channel prediction mask we use squeeze-and-excitation module [31] (SE module) to study the important weights of different channels which will enhance important and restrain unimportant channel features automatically. Then use the Conv2-2 which preserves better edge information to guide the Convi-3, i∈[3,6]. Finally, after fuse the O2OGM feature maps to use cross-entropy loss to compute the final sum loss between PSFEM and O2OGM. More details of these modules can be found in original EGNet paper.

### 3.1 Fire Module

In the model space, we explore the convolution structure and compress the size of the model. The purpose is to achieve the compression of the model parameters while maintaining the same receptive field with smaller convolution kernel. The core idea of SqueezeNet design is introducing the fire module. The design architecture is shown in Figure 2, using 1×1 convolution kernel can limit the number of channels and enhance the abstract expression ability of convolution network which is equivalent to using line sense machine. A larger activation graph is provided for convolution by postponing the down sampling operation which retains more information and can provide higher classification accuracy. The design scale is shown in Figure 3.

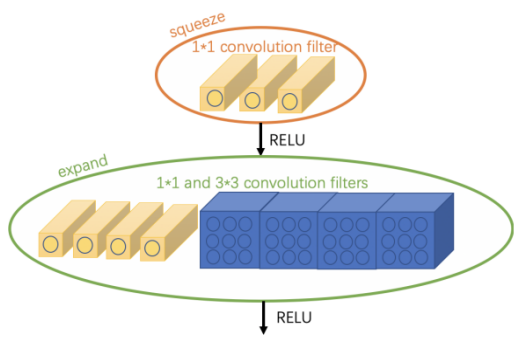

Figure 2. The squeeze and expand Module architecture

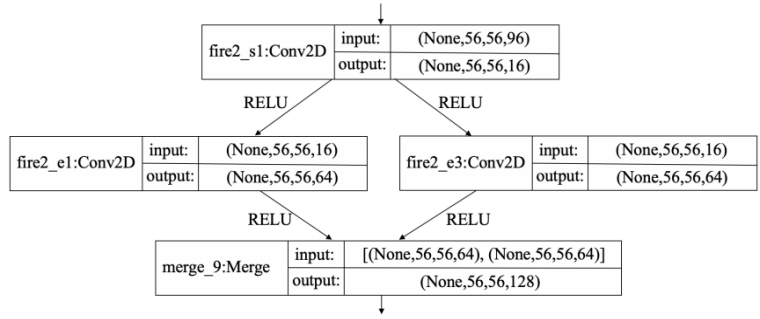

Figure 3. The detailed proportion of squeeze and expand module.

## 3.2 Squeeze-and-Excitation Module

We explore the channel attention mechanism in order to improve the focus of the feature graph after up sampling on the U-shape structure decoder. We introduce the SE module to allocate the available feature resources to the most valuable information in an input semaphore and enhance the expression ability of the network by aggregating the attention weight ratio between different channels.

## 3.3 Our Improved Salient Features Compression Reinforcement Extraction Module

Our improved module (ISFCREM) use 1×1 convolution kernel on the squeeze portion to reduce dimension and combining 3×3 with 1 ×1 convolution kernel of expand portion keep the ratio of 1:1 to increase dimension which not only decreases the kernel parameters caused by the 3 × 3 convolution kernel but also retain the same received sensation. The number of input channels in the fire module keeps the same receptive field as the original convolution to the greatest extent and reduces the count of parameters.

Secondly, concatenate the SE module. SE module first compresses the global spatial information by using the global average pooling to generate the statistical information between channels for a channel descriptor so that the following network layer can obtain the global receptive field information. In the second step, one fully connected layer is used to compress the number of channels to reduce the amount of computation and then the second fully connected layer is used to recover the number of channels. The two fully connected layers leverage the different channel correlation to enhance important features and weaken unnecessary features. Finally, the sigmoid activation function is used to enlarge and shrink the range and the scale factor is fused into the original feature map as the weight ratio then add directly for enhancing the weight of important feature map.

Our improved module (ISFCREM) compress the model while maintaining maximum accuracy.

## 3.4 The Optimizer of AdaX

Compared with the Adam optimizer [32], we use a new optimizer that is AdaX [33] optimizer. The main formulation of updating parameters can be seen as Table 1.

Table 1. The different formulation of optimization strategy between Adam and AdaX optimizer

| $g_t = \nabla_\theta L(\theta_t)$ | (1) | $g_t = \nabla_\theta L(\theta_t)$ | (7) |
|---|---|---|---|
| $m_t = \beta_1 m_{t-1} + (1-\beta_1)g_t$ | (2) | $m_t = \beta_1 m_{t-1} + (1-\beta_1)g_t$ | (8) |
| $v_t = \beta_2 v_{t-1} + (1-\beta_2)g_t^2$ | (3) | $v_t = (1+\beta_2)v_{t-1} + \beta_2 g_t^2$ | (9) |
| $\hat{m}_t = m_t/(1-\beta_1^t)$ | (4) | $\hat{v}_t = v_t/((1+\beta_2)^t - 1)$ | (10) |
| $\hat{v}_t = v_t/(1-\beta_2^t)$ | (5) | $\theta_t = \theta_{t-1} - \alpha_t m_t/\sqrt{\hat{v}_t + \epsilon}$ | (11) |
| $\theta_t = \theta_{t-1} - \alpha_t \hat{m}_t/\sqrt{\hat{v}_t + \epsilon}$ | (6) | | |
| (a) The mainly updating formula (1-6) of AdaX optimizer | | (b) The mainly updating formula (7-11) of Adam optimizer | |

The difference of the two optimizer is that AdaX optimizer remove the offset correction of momentum (4) firstly. Secondly, AdaX optimizer can adapt its learning rate by itself with exponential longterm memory. As time goes on, the decay strategy of Adam optimizer learning rate may not converge a constant but we hope that the learning rate to be a constant because the gradient becomes smaller and the training itself tends to be stable in the later stage of training. So the significance of correcting the learning rate is not great and the correction strength of the leaning rate should become smaller that should be better a constant. As the formulation can be seen that AdaX optimizer can better convergence a constant that make the loss convergence smoother than Adam optimizer. Our training experiment results using three different optimizer for a comparison can be shown as Figure 4.

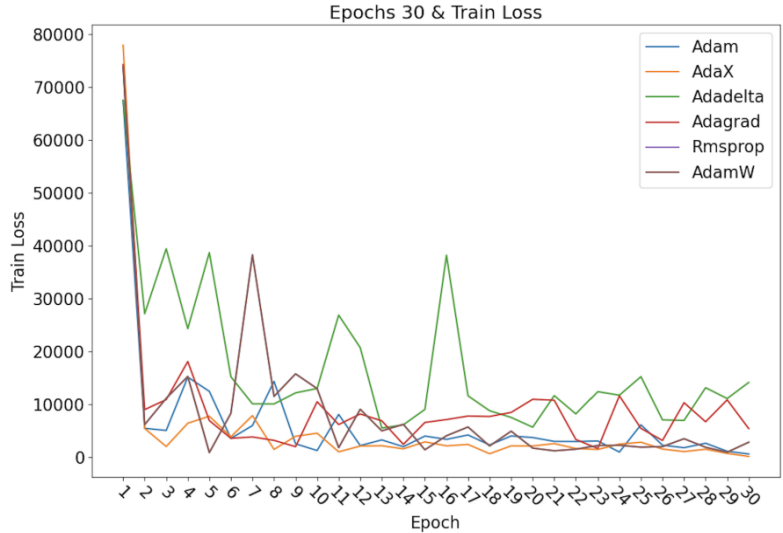

Figure 4. The comparison between Adadelta, Adam, Adagrad, Rmsprop, AdamW and AdaX optimizer based on Resnet-50 as backbone after training 30 epochs on DUTS-TE salient object detection dataset.

## 4 EXPERIMENTS

### 4.1 Implementation Details

We train the improved model on DUTS-TS [34] dataset in our device which contains two P40 GPUs having 24GB video random access memory each other. We also use VGG-16 and Resnet-50 [35] as backbone for a comparison respectively. Our code is implied by PyTorch. We set the weights of newly added convolution layers are initialized randomly with a truncated normal (σ=0.01), and the biases are initialized to 0. We use AdaX as the optimizer. The hyperparameters are set as followed: learning rate=5e-5, weight decay=0.0005, momentum = 0.9, loss weight for each side output is equal to 1. A back propagation is processing for each of the ten images. We do not use the validation dataset during training. We train our model 18 epochs and divide the learning rate by 10 after 9 epochs. We apply the official model to train original experiment as comparison. The results can be shown as Figure 5.

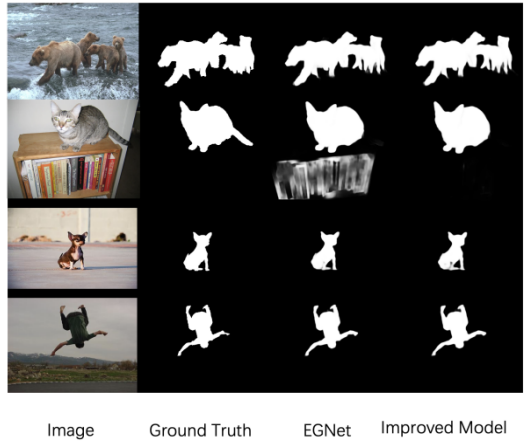

Figure 5. The result salient map of improved model based on the Resnet-50 can be seen as a contrast with the original net. As we can see, the salient object can be focus more of its construction when we imported the SE module in the second line feature maps.

**4.2 Datasets and Evaluation Criteria**

We test the improved model on six widely public benchmark datasets: DUTOMRON [36], DUTS-TE, ECSSD, PASCAL-S [37], HKU-IS [38] and SOD [39] that altogether nearly contains 30000 meaningful semantic images with various complex scenes. DUTS dataset contains 10553 original images, saliency images and corresponding edge images. DUTOMRON dataset contains 5168 original images and salient feature images respectively. Images in this dataset contain one or more salient objects with a relatively complex background. SOD, ECSSD, PASCAL-S and HKU-IS dataset includes 300, 1000, 850, 4447 original images and salient feature images respectively. DUTS-TE dataset includes 5017 original images and salient feature images respectively. The dataset of DUTS is the largest salient object detection stand which include more challenged images with various locations and scales.

We use four metrics, F-Value, mean absolute error (MAE), S-measure [40] to compare with baseline model. F-value is a mean of average precision and average recall, the formula can be described as Equation (12):

$$F_\beta = \frac{(1+\beta^2)*Precision*Recall}{\beta^2*Precision+Recall} \quad (12)$$

where $\beta^2$ is set to 0.3 as done in previous work to weight precision more than recall. The MAE score indicates how similar a saliency map S is compared to the ground truth G and the formula as Equation (13):

$$MAE = \frac{1}{W*H}\sum_{x=1}^{W}\sum_{y=1}^{H}|S(x,y) - G(x,y)| \quad (13)$$

where W and H denote the width and height of S, respectively. The IOU can be described as the intersection-over-union between the predicted map S and ground truth G and the formula as Equation (14):

$$IOU = \frac{S \cap G}{S \cup G} \quad (14)$$

S-measure focuses on evaluating the structural information of salient maps. It is closer to the human visual system than F-measure. We include S-measure for a more comprehensive evaluation for comparison. S-measure could be computed as Equation (15):

$$S = \gamma S_o + (1-\gamma)S_r \quad (15)$$

## 4.3 Evaluation Results

First of all, we compare the traditional mainstream method with our improved model on ESSCD, PASCAL-S, DUTOMRON, HKU-IS, SOD and DUTS-TE databases respectively, and the comparison results are shown in Table 2. We draw the PR curve on DUTOMRON, PASCAL-S, and DUTS-TE datasets which are shown on Figure 6. Next, we also compared the results of IOU, Inference Speed between EGNet and our improved model. The result shown as Table 3 and Table 4 respectively.

Table 2. Quantitative comparison including max F-measure, MAE, and S-measure over six widely used datasets. "-" denotes that corresponding methods are trained on that dataset. "*" means methods using pre-processing or post-processing. Compared with traditional methods that our improved model evaluation result is nearly the same as EGNet but more better than others results. The best results are marked in red and the second best results are marked blue respectively.

|  | ECSSD | | | PASCAL-S | | | DUTOMRON | | | HKU-IS | | | SOD | | | DUTS-TE | | |
|---|---|---|---|---|---|---|---|---|---|---|---|---|---|---|---|---|---|---|
|  | MaxF | MAE | S | MaxF | MAE | S | MaxF | MAE | S | MaxF | MAE | S | MaxF | MAE | S | MaxF | MAE | S |
| VGG-based | | | | | | | | | | | | | | | | | | |
| DCL* | 0.898 | 0.082 | 0.863 | 0.806 | 0.114 | 0.790 | 0.735 | 0.094 | 0.742 | 0.895 | 0.061 | 0.860 | 0.831 | 0.133 | 0.749 | 0.784 | 0.080 | 0.785 |
| DSS* | 0.905 | 0.066 | 0.881 | 0.822 | 0.101 | 0.795 | 0.760 | 0.073 | 0.766 | 0.901 | 0.052 | 0.879 | 0.835 | 0.125 | 0.745 | 0.812 | 0.066 | 0.814 |
| MSR | 0.904 | 0.059 | 0.875 | 0.840 | 0.082 | 0.801 | 0.791 | 0.072 | 0.767 | 0.909 | 0.045 | 0.853 | 0.840 | 0.109 | 0.757 | 0.825 | 0.064 | 0.810 |
| NLDF | 0.905 | 0.066 | 0.874 | 0.821 | 0.098 | 0.805 | 0.752 | 0.080 | 0.751 | 0.900 | 0.049 | 0.878 | 0.837 | 0.125 | 0.755 | 0.814 | 0.067 | 0.807 |
| RAS [41] | 0.917 | 0.060 | 0.885 | 0.832 | 0.104 | 0.799 | 0.784 | 0.062 | 0.793 | 0.911 | 0.048 | 0.886 | 0.845 | 0.132 | 0.762 | 0.802 | 0.060 | 0.826 |
| ELD* | 0.866 | 0.084 | 0.840 | 0.772 | 0.123 | 0.756 | 0.736 | 0.093 | 0.744 | 0.845 | 0.073 | 0.825 | 0.761 | 0.154 | 0.706 | 0.745 | 0.093 | 0.750 |
| DHS [42] | 0.904 | 0.064 | 0.885 | 0.826 | 0.090 | 0.807 | - | - | - | 0.891 | 0.052 | 0.870 | 0.825 | 0.129 | 0.749 | 0.814 | 0.064 | 0.809 |
| RFCN*[43] | 0.897 | 0.096 | 0.854 | 0.825 | 0.120 | 0.799 | 0.747 | 0.096 | 0.753 | 0.894 | 0.080 | 0.860 | 0.807 | 0.163 | 0.731 | 0.784 | 0.088 | 0.782 |
| UCF [44] | 0.906 | 0.082 | 0.886 | 0.820 | 0.126 | 0.804 | 0.732 | 0.133 | 0.749 | 0.886 | 0.075 | 0.872 | 0.799 | 0.164 | 0.760 | 0.770 | 0.114 | 0.778 |
| Amulet | 0.913 | 0.063 | 0.895 | 0.827 | 0.094 | 0.820 | 0.735 | 0.085 | 0.771 | 0.887 | 0.050 | 0.888 | 0.797 | 0.148 | 0.753 | 0.773 | 0.076 | 0.798 |
| C2S [45] | 0.910 | 0.057 | 0.892 | 0.843 | 0.082 | 0.839 | 0.760 | 0.074 | 0.785 | 0.899 | 0.045 | 0.887 | 0.823 | 0.124 | 0.765 | 0.811 | 0.060 | 0.821 |
| PAGR | 0.924 | 0.062 | 0.888 | 0.849 | 0.088 | 0.817 | 0.772 | 0.072 | 0.752 | 0.917 | 0.047 | 0.887 | 0.842 | 0.146 | 0.718 | 0.853 | 0.055 | 0.824 |
| EGNet | 0.942 | 0.045 | 0.914 | 0.863 | 0.075 | 0.846 | 0.824 | 0.054 | 0.813 | 0.927 | 0.036 | 0.912 | 0.870 | 0.111 | 0.789 | 0.880 | 0.045 | 0.868 |
| Ours | 0.936 | 0.045 | 0.909 | 0.855 | 0.081 | 0.837 | 0.815 | 0.060 | 0.815 | 0.925 | 0.037 | 0.903 | 0.854 | 0.114 | 0.778 | 0.869 | 0.046 | 0.865 |
| ResNet-based | | | | | | | | | | | | | | | | | | |
| SRM*[49] | 0.914 | 0.056 | 0.897 | 0.836 | 0.085 | 0.832 | 0.770 | 0.071 | 0.775 | 0.908 | 0.044 | 0.888 | 0.842 | 0.128 | 0.743 | 0.824 | 0.059 | 0.824 |
| DGRL[52] | 0.922 | 0.044 | 0.908 | 0.844 | 0.074 | 0.840 | 0.775 | 0.062 | 0.793 | 0.912 | 0.038 | 0.898 | 0.845 | 0.105 | 0.772 | 0.828 | 0.050 | 0.837 |
| PiCANet*[33] | 0.934 | 0.046 | 0.914 | 0.865 | 0.078 | 0.851 | 0.822 | 0.062 | 0.809 | 0.922 | 0.045 | 0.906 | 0.861 | 0.101 | 0.788 | 0.862 | 0.051 | 0.850 |
| EGNet | 0.944 | 0.042 | 0.917 | 0.870 | 0.074 | 0.853 | 0.843 | 0.053 | 0.817 | 0.936 | 0.031 | 0.919 | 0.892 | 0.097 | 0.806 | 0.892 | 0.040 | 0.876 |
| Ours | 0.938 | 0.043 | 0.913 | 0.858 | 0.074 | 0.840 | 0.818 | 0.058 | 0.815 | 0.928 | 0.035 | 0.905 | 0.865 | 0.116 | 0.774 | 0.858 | 0.044 | 0.867 |

Table 3. The evaluation of IOU between different datasets compared EGNet with our improved model.

|  | ECSSD | PASCAL-S | DUTOMRON | HKU-IS | SOD | DUTS-TE |
|---|---|---|---|---|---|---|
| VGG-based | | | | | | |
| **EGNet** | 0.867 | 0.750 | 0.697 | 0.843 | 0.654 | 0.767 |
| **Ours** | 0.846 | 0.728 | 0.658 | 0.828 | 0.639 | 0.737 |
| Resnet-based | | | | | | |
| **EGNet** | 0.878 | 0.755 | 0.701 | 0.858 | 0.692 | 0.782 |
| **Ours** | 0.854 | 0.735 | 0.654 | 0.832 | 0.637 | 0.740 |

Table 4. The evaluation of inference speed between different datasets compared EGNet with our improved model. The inference speed of our improved model is faster than EGNet.

|  | PASCAL-S (seconds/sheet) | DUTOMRON (seconds/sheet) | HKU-IS (seconds/sheet) | SOD (seconds/sheet) | DUTS-TE (seconds/sheet) |
|---|---|---|---|---|---|
| VGG-based | | | | | |
| **EGNet** | 0.476 | 0.126 | 0.244 | 0.117 | 0.113 |
| **Ours** | 0.448 | 0.142 | 0.223 | 0.090 | 0.122 |
| Resnet-based | | | | | |
| **EGNet** | 0.128 | 0.131 | 0.270 | 0.180 | 0.129 |
| **Ours** | 0.113 | 0.076 | 0.198 | 0.103 | 0.078 |

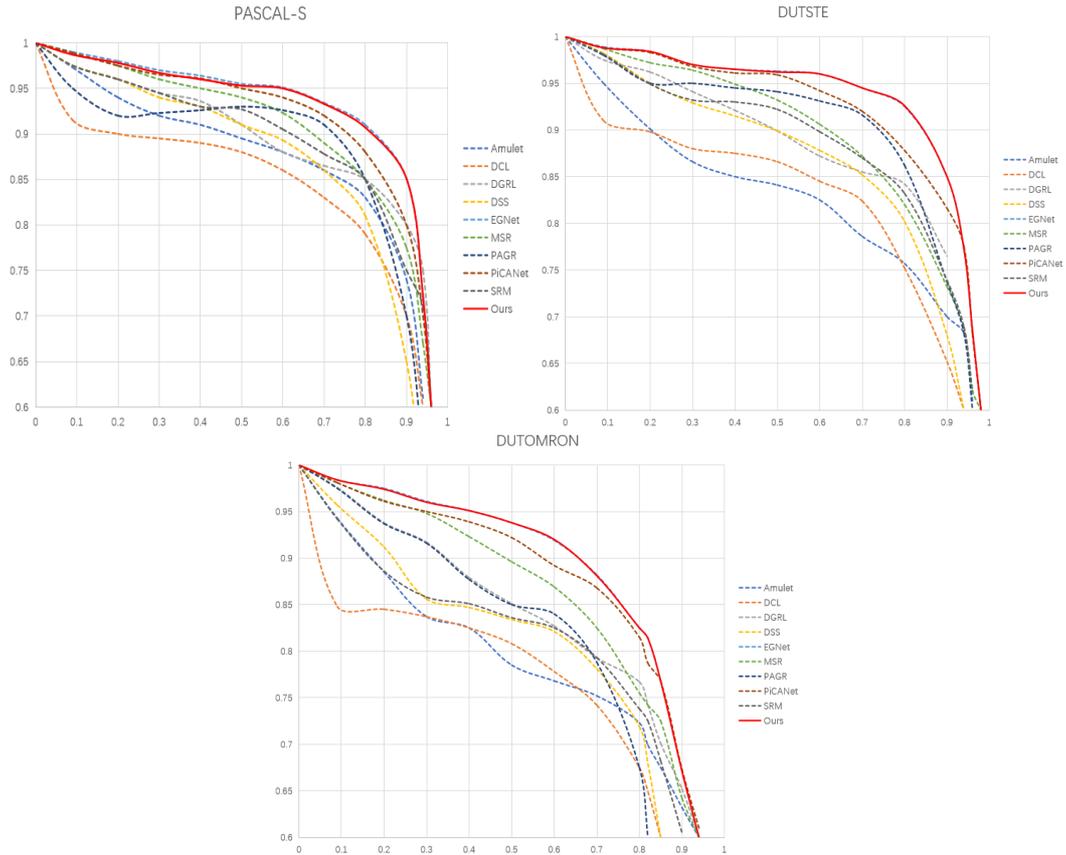

Figure 6. The Precision and Recall for PASCAL-S, DUTSTE, DUTOMRON datasets respectively. Our improved model is better than others and nearly similar as the state-of-the-art EGNet.

## 5. CONCLUSIONS

In this paper, we use smaller CNNs compressed the model size based on the EGNet structure firstly. The parameters are reduced from 108068426 to 30175602 (111692618 to 33799794) and the model size from 432MB to 115MB (426MB to 129MB) using VGG-16 (Resnet-50) as the backbone respectively. It is a third of the original size. But we found that the accuracy is lower 3.0+ compared with the-state-of-art that is EGNet. For a remedy, we introduce channel attention to minimize gaps. The results shown that our improved model is close to the-state-of-the-art. We also explore a new optimizer for better and faster convergence a global minimum loss. The experiment shown that its test result is crowded on the Adam optimizer which is the best optimizer widely used but nearly half of training time.

In the future work, we want to explore a larger proportion of compression models by means of model weight pruning and knowledge distillation. In terms of model accuracy, we will continue to explore the edge structure extraction of refined feature map to make the segmented region more effective. At the same time, we will explore new self-attention structure mechanisms, such as introducing the latest transformer mechanism to capture global context information and improve segmentation accuracy.

### ACKNOWLEDGEMENTS

This work is supported by CERNET Innovation Project (NGII20190625) and The Inner Mongolia Natural Science Foundation of China under Grant No.2021MS06016.

# REFERENCES

[1] L SeunghoonHong, TackgeunYou, SuhaKwak, and Bohyung Han. Online tracking by learning discriminative saliency map with convolutional neural network. In ICML, pages 597–606, 2015. 1

[2] Ming-Ming Cheng, Fang-Lue Zhang, Niloy J Mitra, Xiaolei Huang, and Shi-Min Hu. Repfinder: finding approximately repeated scene elements for image editing. ACM TOG, 29(4):83, 2010. 1

[3] Celine Craye, David Filliat, and Jean-Francois Goudou. Environment exploration for objectbased visual saliency learning. In ICRA, pages 2303–2309, 2016. 1

[4] Laurent Itti, Christof Koch, and Ernst Niebur. A model of saliencybased visual attention for rapid scene analysis. IEEE TPAMI, 20(11):1254–1259, 1998. 1

[5] Tie Liu, Zejian Yuan, Jian Sun, Jingdong Wang, Nanning Zheng, Xiaoou Tang, and Heung-Yeung Shum. Learning to detect a salient object. IEEE TPAMI, 33(2):353–367, 2011. 1, 5, 6, 7

[6] Dominik A Klein and Simone Frintrop. Center-surround divergence of feature statistics for salient object detection. In ICCV, 2011. 1

[7] Federico Perazzi, Philipp Kra ̈henbu ̈hl, Yael Pritch, and Alexander Hornung. Saliency filters: Contrast based filtering for salient region detection. In CVPR, pages 733–740, 2012. 1, 2

[8] Ali Borji and Laurent Itti. Exploiting local and global patch rarities for saliency detection. In CVPR, pages 478–485, 2012. 1

[9] Huaizu Jiang, Jingdong Wang, Zejian Yuan, Yang Wu, Nanning Zheng, and Shipeng Li. Salient object detection: A discriminative regional feature integration approach. In CVPR, pages 2083–2090, 2013. 1, 2

[10] Qiong Yan, Li Xu, Jianping Shi, and Jiaya Jia. Hierarchical saliency detection. In CVPR, pages 1155–1162, 2013. 1, 5, 8

[11] Ming Cheng, Niloy J Mitra, Xumin Huang, Philip HS Torr, and Song Hu. Global contrast based salient region detection. IEEE TPAMI, 2015. 1, 2, 8

[12] Qibin Hou, Ming-Ming Cheng, Xiaowei Hu, Ali Borji, Zhuowen Tu, and Philip Torr. Deeply supervised salient object detection with short connections. IEEE TPAMI, 41(4):815–828, 2019. 1, 2, 3, 5, 6, 7, 8

[13] Zhiming Luo, Akshaya Kumar Mishra, Andrew Achkar, Justin A Eichel, Shaozi Li, and Pierre-Marc Jodoin. Nonlocal deep features for salient object detection. In CVPR, 2017. 1, 2, 7, 8

[14] Pingping Zhang, Dong Wang, Huchuan Lu, Hongyu Wang, and Xiang Ruan. Amulet: Aggregating multilevel convolutional features for salient object detection. In ICCV, 2017. 1, 2, 6, 7, 8

[15] Olaf Ronneberger, Philipp Fischer, and Thomas Brox. U-net: Convolutional networks for biomedical image segmentation. In International Conference on Medical image computing and computer-assisted intervention, pages 234–241, 2015. 1

[16] Tsung-Yi Lin, Piotr Dolla ́r, Ross B Girshick, Kaiming He, Bharath Hariharan, and Serge J Belongie. Feature pyramid networks for object detection. In CVPR, 2017. 1, 2, 3, 4, 6

[17] Guanbin Li, Yuan Xie, Liang Lin, and Yizhou Yu. Instance-level salient object segmentation. In CVPR, 2017. 1, 5, 6, 7, 8

[18] TiantianWang, LiheZhang, ShuoWang, HuchuanLu, Gang Yang, Xiang Ruan, and Ali Borji. Detect globally, refine locally: A novel approach to saliency detection. In CVPR, pages 3127–3135, 2018. 1, 3, 6, 7, 8

[19] ibin Hou, Jiang-Jiang Liu, Ming-Ming Cheng, Ali Borji, and Philip HS Torr. Three birds one stone: A unified framework for salient object segmentation, edge detection and skeleton extraction. arXiv preprint arXiv:1803.09860, 2018. 1, 7, 8

[20] Zhao J, JJ Liu, Fan D P, et al. EGNet: Edge Guidance Network for Salient Object Detection[C]// 2019 IEEE/CVF International Conference on Computer Vision (ICCV). IEEE, 2020.

[21] Iandola F N, Han S, Moskewicz M W, et al. SqueezeNet: AlexNet-level accuracy with 50x fewer parameters and <0.5MB model size[J]. 2016.

[22] Xiaohui Li, Huchuan Lu, Lihe Zhang, Xiang Ruan, and Ming Hsuan Yang. Saliency detection via dense and sparse reconstruction. In ICCV, pages 2976–2983, 2013. 2

[23] LEE G, TAI Y W, KIM J, et al. Deep saliency with encoded low level distance map and high level features[C]//Proceedings of the 2016 IEEE Conference on Computer Vision and Pattern Recognition, Las Vegas, Jun 27-30, 2016. Washington: IEEE Computer Society, 2016: 660-668.

[24] WANG T T, ZHANG L H, LU H C, et al. Kernelized subspace ranking for saliency detection[C]//LNCS 9912: Proceedings of the 14th European Conference on Computer Vision, Amsterdam, Oct 11- 14, 2016. Berlin, Heidelberg: Springer, 2016: 450-466.

[25] LI G B, YU Y Z. Deep contrast learning for salient object detection[C]//Proceedings of the 2016 IEEE Conference on Computer Vision and Pattern Recognition, Las Vegas, Jun 27- 30, 2016. Washington: IEEE Computer Society, 2016: 478-487.

[26] DAI J F, LI Y, HE K, et al. R-FCN: object detection via region- based fully convolutional networks[C]//Proceedings of the 30th International Conference on Neural Information Processing Systems, Barcelona, Dec 5, 2016. Massachusetts: MIT Press, 2016: 379-387.

[27] WU Z, SU L, HUANG Q M, et al. Cascaded partial decoder for fast and accurate salient object detection[C]//Proceedings of the 2019 IEEE Conference on Computer Vision and Pat- tern Recognition, Long Beach, Jun 16-20, 2019. Washington: IEEE Computer Society, 2019: 3907-3916.

[28] LIU J J, HOU Q B, CHENG M M, et al. A simple pooling-based design for real-time salient object detection[C]//Proceedings of the 2019 IEEE/CVF Conference on Computer Vision and Pattern Recognition, Long Beach, Jun 16-20, 2019. Piscataway: IEEE, 2019: 3917-3926.

[29] ZHANG L, DAI J, LU H C, et al. A bi-directional message passing model for salient object detection[C]//Proceedings of the 2018 IEEE Conference on Computer Vision and Pat- tern Recognition, Salt Lake City, Jun 18-22, 2018. Washing- ton: IEEE Computer Society, 2018: 1741-1750.

[30] Simonyan K, Zisserman A. Very Deep Convolutional Networks for Large-Scale Image Recognition[J]. Computer Science, 2014.

[31] Jie H, Li S, Gang S. Squeeze-and-Excitation Networks[C]// 2018 IEEE/CVF Conference on Computer Vision and Pattern Recognition (CVPR). IEEE, 2018.

[32] Kingma D, Ba J. Adam: A Method for Stochastic Optimization[J]. Computer Science, 2014.

[33] Li W, Zhang Z, Wang X, et al. AdaX: Adaptive Gradient Descent with Exponential Long Term Memory[J]. 2020.

[34] Xiaoning Zhang, Tiantian Wang, Jinqing Qi, Huchuan Lu, and Gang Wang. Progressive attention guided recurrent network for salient object detection. In *CVPR*, pages 714–722, 2018. 1, 3, 6, 7, 8

[35] He K, Zhang X, Ren S, et al. Deep Residual Learning for Image Recognition[C]// 2016 IEEE Conference on Computer Vision and Pattern Recognition (CVPR). IEEE, 2016.

[36] Chuan Yang, Lihe Zhang, Huchuan Lu, Xiang Ruan, and Ming-Hsuan Yang. Saliency detection via graph-based manifold ranking. In *CVPR*, pages 3166–3173, 2013.

[37] Yin Li, Xiaodi Hou, Christof Koch, James M Rehg, and Alan L Yuille. The secrets of salient object segmentation. In *CVPR*, pages 280–287, 2014.

[38] Guanbin Li and Yizhou Yu. Visual saliency based on multi- scale deep features. In *CVPR*, pages 5455–5463, 2015.

[39] David Martin, Charless Fowlkes, Doron Tal, and Jitendra Malik. A database of human segmented natural images and its application to evaluating segmentation algorithms and measuring ecological statistics. In *ICCV*, volume 2, pages 416–423, 2001.

[40] Cheng M M, Fan D P. Structure-Measure: A New Way to Evaluate Foreground Maps[J]. International Journal of Computer Vision, 2021, 129(9):2622-2638.

[41] Shuhan Chen, Xiuli Tan, Ben Wang, and Xuelong Hu. Reverse attention for salient object detection. In *ECCV*, pages 234–250, 2018.

[42] Nian Liu and Junwei Han. Dhsnet: Deep hierarchical saliency network for salient object detection. In *CVPR*, pages 678–686, 2016.

[43] Linzhao Wang, Lijun Wang, Huchuan Lu, Pingping Zhang, and Xiang Ruan. Saliency detection with recurrent fully convolutional networks. In *ECCV*, 2016.

[44] Pingping Zhang, Dong Wang, Huchuan Lu, Hongyu Wang, and Baocai Yin. Learning uncertain convolutional features for accurate saliency detection. In *ICCV*, pages 212–221. IEEE, 2017.

[45] Xin Li, Fan Yang, Hong Cheng, Wei Liu, and Dinggang Shen. Contour knowledge transfer for salient object detection. In *ECCV*, pages 355–370, 2018.

**Authors**

**Ming Ma** is a lecturer of Inner Mongolia University, graduated from Pai Chai University in Korea with a PhD in 2009. He joined the School of Computer Science of Inner Mongolia University as a teacher from 2013.

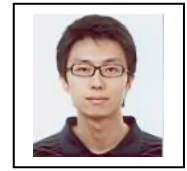

**Bin Zhang** is currently a master of computer technology, Inner Mongolia University, under the super-vision of Lecturer Ming Ma. His research interests include deep learning, image processing, and computer vision.

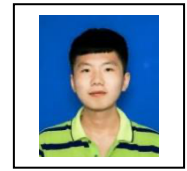

**Yang Wu** is a doctoral student at the College of Computer Science of Inner Mongolia University. He is mainly interested in action recognition and video classification tasks.

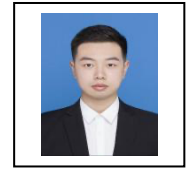

**Xiaojing Zhang** is currently a graduate student with College of Electronic Information Technology, Inner Mongolia University, under the super-vision of Lecturer Ming Ma. Her research interests include deep learning, image processing, and computer vision.

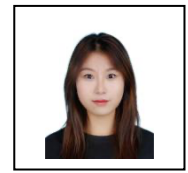